\def\year{2021}\relax
\documentclass[letterpaper]{article} %
\usepackage{aaai21}  %
\usepackage{times}  %
\usepackage{helvet} %
\usepackage{courier}  %
\usepackage[hyphens]{url}  %
\usepackage{graphicx} %
\usepackage{natbib}  %
\usepackage{caption} %
\usepackage{comment}
\usepackage{xcolor}
\usepackage{algorithm}
\usepackage{algpseudocode} %
\usepackage{url}             %
\usepackage{multirow}
\usepackage{cite}
\usepackage{subcaption}
\usepackage{booktabs}

\usepackage{latexsym}
\usepackage{amsmath}
\usepackage{amssymb} %
\usepackage{mathtools}       %
\usepackage{stmaryrd}        %
\usepackage{bm}              %

\usepackage{amsthm}
\usepackage{cleveref}

\providecommand{\argmax}{\operatornamewithlimits{argmax}} %
\DeclareMathOperator{\Var}{Var}   %
\DeclareMathOperator{\Cov}{Cov}   %
\providecommand{\R}{\mathbb{R}} %
\providecommand{\E}{\mathbb{E}} %
\providecommand{\T}{\mathrm{T}} %
\DeclarePairedDelimiterX{\inner}[2]{\langle}{\rangle}{#1, #2}
\DeclarePairedDelimiter{\norm}{\lVert}{\rVert}
\DeclarePairedDelimiter{\abs}{\lvert}{\rvert}

\newtheorem{theorem}{Theorem}[section]

\newcommand{\op}{\mathcal{O}}
\newcommand{\av}{a}

\newcommand{\blue}[1]{\textcolor{blue}{}}

\title{AdvantageNAS: Efficient Neural Architecture Search with Credit Assignment}
\author {
        Rei Sato,
        Jun Sakuma,
        Youhei Akimoto \\
}
\affiliations {
    University of Tsukuba \\
    RIKEN AIP \\
    \{reisato@bbo., jun@, akimoto@\}cs.tsukuba.ac.jp
}
\begin{document}

\maketitle

\begin{abstract}
Neural architecture search (NAS) is an approach for automatically designing a neural network architecture without human effort or expert knowledge. However, the high computational cost of NAS limits its use in commercial applications. Two recent NAS paradigms, namely one-shot and sparse propagation, which reduce the time and space complexities, respectively, provide clues for solving this problem. In this paper, we propose a novel search strategy for one-shot and sparse propagation NAS, namely AdvantageNAS, which further reduces the time complexity of NAS by reducing the number of search iterations. AdvantageNAS is a gradient-based approach that improves the search efficiency by introducing credit assignment in gradient estimation for architecture updates. Experiments on the NAS-Bench-201 and PTB dataset show that AdvantageNAS discovers an architecture with higher performance under a limited time budget compared to existing sparse propagation NAS. To further reveal the reliabilities of AdvantageNAS, we investigate it theoretically and find that it monotonically improves the expected loss and thus converges.
\end{abstract}

\section{Introduction}

Deep neural networks (DNNs) have achieved remarkable performance improvements in various domains.
Finding a suitable DNN architecture is critical for applying a DNN to a new task and achieving higher performance.
However, as the architecture is defined by a combination of tens to hundreds of hyperparameters (filter sizes of convolutional layers, existence of connections between some layers, etc.), improving such a combination requires considerable human effort and expert knowledge.
Neural architecture search (NAS) has been studied as a promising approach for automatically identifying sophisticated DNN architectures and overcoming the problems in applying DNNs to previously unseen tasks.

However, the high computational cost (time and space complexity) of NAS is a bottleneck in its applications, especially commercial applications that do not involve multiple high-spec GPUs.
Hence, reducing the computational cost of NAS is an important research direction.

One-shot NAS reduces the time complexity of NAS by simultaneously optimizing the architecture and the parameters (weights) in a single training process or a few training processes.
It maintains a redundant deep learning model comprising numerous layers and connections, called the parent network, which can represent all possible architectures in the search space as its sub-network.
All the parameters required by candidate architectures are possessed by the parent network and shared among different architectures.
The goodness of an architecture is evaluated using the shared parameters, which drastically reduces the computational time for the architecture search compared to conventional approaches that require the parameters to be trained from scratch for each candidate architecture evaluation.

From the viewpoint of GPU memory consumption, one-shot NAS can be categorized into dense propagation NAS (DP-NAS) and sparse propagation NAS (SP-NAS).
DP-NAS requires the entire parent network to be loaded in the GPU memory to perform forward and/or backward propagation, whereas SP-NAS requires only a sub-network to be loaded.
DP-NAS is more popular owing to the success of DARTS and its successors \cite{liu2018darts,xie2018snas,yao2020nasp}.
However, because of its high memory consumption, its search space is limited.
SP-NAS overcomes this limitation and provides more flexible search spaces on a standard GPU with limited memory.
Hence, we focus on SP-NAS in this study.

To further reduce the computational cost of one-shot SP-NAS, we propose a novel one-shot SP-NAS, namely \emph{AdvantageNAS}.
Although various factors affect the performance of NAS (performance estimation, search space, etc.) and state-of-the-art (SOTA) performance may require a search for the best combination of these factors, this study focuses on the search strategy.
We emphasize that the objective of this study is not to obtain the SOTA architecture but to investigate the efficiency of the search strategy.

AdvantageNAS is a gradient-based SP-NAS.\footnote{The advantages of gradient-based approaches are as follows. Approaches using evolutionary algorithms (e.g., \cite{real2017evo}) or progressive search methods (e.g., \cite{liu2018pnas}) must define a mutation or expansion strategy depending on the search space. Such manual preparations depending on the search space are not required in gradient-based methods. Another advantage of gradient-based methods is that they can be easily implemented in most modern deep learning frameworks, as they support automatic differentiation of deep neural networks.}
This category of NAS includes ENAS \cite{pham2018enas}, ProxylessNAS \cite{cai2018proxylessnas}, PARSEC \cite{casale2019parsec}, and GDAS \cite{xuanyi2019gdas}.
These methods introduce a family of parametric probability distributions from which an architecture is sampled, and the distribution parameter is updated with a gradient of the objective function estimated using the Monte Carlo method, which indirectly optimizes the architecture itself.
As the gradient is estimated, reducing the variance of the gradient estimate intuitively improves the search efficacy. 
Inspired by the actor-critic approach \cite{konda2000ac}, we adopt the notion of \emph{advantage}, which has been introduced to solve the well-known \emph{credit assignment} problem of policy gradients \cite{Minsky:1961:ire}.

We investigate AdvantageNAS both theoretically and empirically.
Empirical studies show that it performs well on different tasks with different search spaces, whereas theoretical studies guarantee its performance on empirically unforeseen tasks.
Theoretical studies are relatively scarce in NAS research; however, they are important for investigating search strategies for NAS.

Our contributions are summarized as follows:
\begin{itemize}
\item
  We propose a novel SP-NAS, namely AdvantageNAS.
  Our analysis of a test scenario demonstrates that the estimation variance of the gradient is reduced compared to REINFORCE \cite{Williams1992}, which is a baseline search strategy for ENAS.
\item
  The theoretical investigation reveals that the estimated gradient in AdvantageNAS is an unbiased estimator of the gradient of the expected loss.
  From the unbiasedness, we find that AdvantageNAS monotonically improves the expected loss and thus converges.
  Further, we confirm that reducing the gradient variance contributes to an improved lower bound of the loss improvement in each iteration.
\item
  We compared AdvantageNAS with various gradient-based NAS methods in three different settings to demonstrate its efficacy on different search spaces.
  Our experiments were performed on the same code base to compare the search strategies.\footnote{NAS approaches using different components such as search spaces and training protocols have been compared in the literature. However, it has been reported, e.g.,\ in \cite{Yang2020iclr}, that such differences in the components make it difficult to fairly compare the contributions of each component.
  Our code: \url{https://github.com/madoibito80/advantage-nas}.}
  The results showed that AdvantageNAS converges faster than existing SP-NAS approaches.
\end{itemize}

\section{One-shot Neural Architecture Search}

\subsubsection{DAG Representation of DNNs}
We formulate a deep neural network as a directed acyclic graph (DAG) following previous studies \cite{pham2018enas,liu2018darts}.
In the DAG representation, each node corresponds to a feature tensor $x$ and each edge corresponds to an operation $\op$ applied to the incoming feature tensor, e.g., convolution or pooling. 
The adjacency matrix of the DAG is pre-defined. Let $\mathcal{I}$ be a set of edges.
The input and output nodes are two special nodes; the feature tensor at the input node is the input to the DNN and the input node has no incoming edge, while the feature tensor at the output node is the output of the DNN and the output node has no outgoing edge.

The feature tensor $x_\ell$ at node $\ell$, except for the input node, is computed as follows.
Let $\mathcal{E}_\ell \subseteq \mathcal{I}$ be a subset of the edges that point to node $\ell$.
Let $Q(i)$ be the input node of edge $i \in \mathcal{I}$.
For each edge $i \in \mathcal{E}_\ell$, the operation $\mathcal{O}_i$ is applied to $x_{Q(i)}$ and output $y_i = \mathcal{O}_i(x_{Q(i)})$.
The feature tensor $x_\ell$ is computed by aggregating the outputs $y_i$ of the incoming edges $i \in \mathcal{E}_\ell$ with some pre-defined aggregation function $F_\ell: \{y_i : i \in \mathcal{E}_\ell\} \mapsto x_\ell$ such as summation or concatenation.

\subsubsection{Search Space for Neural Architecture}
For each edge $i \in \mathcal{I}$ of the DAG, a set of candidate operations is prepared.
Let $\mathcal{J}_i$ be the indices of the candidate operations at edge $i$, and each candidate operation is denoted as $\mathcal{O}_i^j$ for $j \in \mathcal{J}_i$.
Each candidate operation may have its own real-valued parameter vector $w_i^j$, often referred to as a \emph{weights}, and $\mathcal{O}_i^j$ is assumed to be differentiable w.r.t.\ the weights.
The operation $\mathcal{O}_i$ of edge $i$ is then defined as a weighted average of the candidate operations with weight $\bm{a}_i = (a_i^1, \dots, a_i^{\abs{\mathcal{J}_i}})^{\T}$:
\begin{equation}
\textstyle  \mathcal{O}_i(x_{Q(i)}) = \sum_{j \in \mathcal{J}_i} \av_{i}^{j} \mathcal{O}_{i}^{j}(x_{Q(i)}; w_{i}^{j}) \enspace,\label{forward}  
\end{equation}
where $\sum_{j \in \mathcal{J}_i} a_i^{j} = 1$. 
The vector $\bm{a} = (\bm{a}_1, \dots, \bm{a}_{\abs{\mathcal{I}}})$ is referred to as the \emph{architecture}. 

The objective of NAS is to locate the best architecture under some performance criterion.
The aim of our study, as well as many related studies \cite{liu2018darts,xie2018snas,cai2018proxylessnas,xuanyi2019gdas,yao2020nasp,casale2019parsec}, is to find the best architecture such that $\bm{a}_i$ for each $i \in \mathcal{I}$ is a one-hot vector.
In other words, we choose one operation at each edge.

\subsubsection{Weight Sharing and One-shot NAS}
One-shot architecture search is a method for jointly optimizing $\bm{a}$ and $\bm{w}$ during a single training process.

Let $R$ be a performance metric (referred to as \emph{reward} in this study) that takes an architecture $\bm{a}$ and weights $\bm{w}_{\bm{a}}$ required to compute the DAG defined by $\bm{a}$. 
The reward is typically the negative expected loss $-\E_{d \sim \mathcal{P}}[\mathcal{L}(\bm{a}, \bm{w}_{\bm{a}}; d)]$, where $\mathcal{L}$ is a data-wise loss, $\mathcal{P}$ is a data distribution, and $d$ represents data.
For example, in the image classification task, $d$ is a set of an image and its label, and $\mathcal{L}$ is the cross-entropy loss using $d$.
The objective of NAS is written as $\max_{\bm{a}} R(\bm{a}, \bm{w}_{\bm{a}}^*)$, where $\bm{w}_{\bm{a}}^* = \argmax_{\bm{w}_{\bm{a}}} R(\bm{a}, \bm{w}_{\bm{a}})$ is the optimal weights for architecture $\bm{a}$.
The expected reward $R$ is often approximated by the average over the validation dataset $\mathcal{V}$ for optimization of $\bm{a}$ and the training dataset $\mathcal{D}$ for optimization of $\bm{w}_{\bm{a}}$.
The difficulty in simultaneously optimizing $\bm{a}$ and $\bm{w}_{\bm{a}}$ is that the domain of $\bm{w}_{\bm{a}}$ changes with $\bm{a}$ during the training process.
The time complexity of conventional NAS approaches (e.g., \cite{zophL2016nasrl}) is extremely high because they require $\bm{w}_{\bm{a}}$ to be trained for each $\bm{a}$ from scratch. 

Weight sharing \cite{pham2018enas} is a technique for simultaneous optimization.
All the weights appearing in the candidate operations are stored in a single weight set, $\bm{w} = \{w_i^{j} : i \in \mathcal{I},\ j \in \mathcal{J}_i \}$. The reward function is extended to accept $\bm{w}$ by defining $R(\bm{a}, \bm{w}) = R(\bm{a}, \bm{w}_{\bm{a}})$.
Then, the objective of NAS is to find $(\bm{a}, \bm{w})$ that maximizes $R(\bm{a}, \bm{w})$.
A disadvantage of the weight sharing approach is that a large parameter set $\bm{w}$ is required during the search.
However, this set can be stored in the CPU memory and only a subset $\bm{w}_{\bm{a}} \subseteq \bm{w}$ is required to be loaded in the GPU, which typically involves much smaller memory.
Moreover, if $\bm{a}$ is sparse during the training, the GPU memory consumption is limited.

\section{Gradient-based One-shot NAS}
\label{sec:related-work}

\begin{table}[t]
\centering
\small
    \begin{tabular}{l|c|c|c|c}
      \hline

      \hline
    Method & Propagation & $\bm{a}_i$ & Stochastic & TG \\
        \hline
    DARTS  & dense & n-hot & no &  \\
    SNAS   & dense & n-hot & yes &  \\
    NASP   & dense\footnotemark & 1-hot &  no & \checkmark \\
    \hline
    ProxylessNAS  & sparse & 2-hot & yes &  \\
    PARSEC        & sparse & 1-hot & yes & \\
    GDAS          & sparse & 1-hot & yes &  \\
    ENAS          & sparse & 1-hot & yes & \\
    \hline
    AdvantageNAS & sparse & 1-hot & yes & \checkmark \\
      \hline

      \hline
    \end{tabular}
\caption{Properties of various gradient-based NAS methods. The ``Propagation'' column indicates whether the method belongs to DP-NAS or SP-NAS discussed in \Cref{sec:related-work}. TG: Theoretical Guarantee.}\label{tab:related}
\end{table}

\footnotetext{NASP performs sparse propagation in the weight update step and dense propagation in the architecture update step.}

The major stream of search strategies for one-shot NAS is gradient-based search \cite{pham2018enas,liu2018darts,xie2018snas,cai2018proxylessnas,xuanyi2019gdas,yao2020nasp,casale2019parsec}. The architecture $\bm{a}$ is either re-parameterized by a real-valued vector $\theta$ through some function $\bm{a} = \mu(\theta)$ or sampled from a probability distribution $\bm{a} \sim p_\theta$ parameterized by $\theta$.
Gradient steps on $\theta$ as well as on $\bm{w}$ are taken alternatively or simultaneously to optimize $\bm{a}$ and $\bm{w}$. 

A comparison of related studies is shown in \Cref{tab:related}.

\subsubsection{Dense Propagation NAS (DP-NAS)}\label{sec:dense}
DP-NAS has been popularized by DARTS \cite{liu2018darts}, where the architecture is re-parameterized by the real-valued vector $\theta \in \mathbb{R}^{\sum_{i\in \mathcal{I}}\abs{\mathcal{J}_i}}$ through the softmax function $\mu(\theta) = (\mu_1(\theta_1), \dots, \mu_{\abs{\mathcal{I}}}(\theta_{\abs{\mathcal{I}}}))$, where $\mu_i(\theta_i) = (\mu_i^1(\theta_1), \dots, \mu_i^{\abs{\mathcal{J}_i}}(\theta_i))$ and 

\begin{equation}
  \textstyle \mu_{i}^{j}(\theta_{i}) = \exp(\theta_{i}^{j}) \big/ \sum_{k \in \mathcal{J}_i}\exp(\theta_{i}^{k}) \enspace. \label{softmax}
\end{equation}
In other words, the output of the $j$th edge in \eqref{forward} is a weighted sum of the outputs of all candidate operations.
The architecture $\bm{a} = \mu(\theta)$ is optimized with the gradient $\nabla_{\theta} R(\mu(\theta), \bm{w})$.

Several follow-up works have been conducted.
SNAS \cite{xie2018snas} uses a concrete distribution \cite{maddison2017concrete} to represent $\bm{a}$.
NASP \cite{yao2020nasp} introduces a proximal iteration to update $\theta$.

\subsubsection{Sparse Propagation NAS (SP-NAS)}\label{sec:sparse}
SP-NAS introduces a parametric probability distribution $p_\theta$ over the architecture space and optimizes $\bm{a}$ indirectly through the optimization of $\theta$.
Thus, all of the SP-NAS is the stochastic algorithm shown in \Cref{tab:related}.
The objective is to maximize the reward function expected over the architecture distribution $J(\theta, \bm{w}) = \E_{\bm{a} \sim p_\theta}[ R(\bm{a}, \bm{w})]$. 
Different distributions and different update rules have been proposed. 
ENAS \cite{pham2018enas} uses LSTM \cite{hoch1997lstm} to represent $p_\theta$ and REINFORCE \cite{Williams1992} to estimate $\nabla_\theta J(\theta, \bm{w})$:
\begin{equation}
  \begin{split}
  \nabla_\theta J(\theta, \bm{w})
  &= \nabla_{\theta} \mathbb{E}_{\bm{a}\sim p_{\theta}(\bm{a})}[R(\bm{a},\bm{w})]
  \\
  &= \mathbb{E}_{\bm{a}\sim p_{\theta}(\bm{a})} [R(\bm{a},\bm{w}) \nabla_{\theta} \log(p_{\theta}(\bm{a}))] \enspace.\label{eq:reinforce}
  \end{split}
\end{equation}
Note that REINFORCE does not need to execute a backward process to update $\theta$ and \eqref{eq:reinforce} can be approximated using the Monte Carlo method with $\bm{a} \sim p_\theta$.

Different follow-up studies have been conducted to accelerate the optimization of $\theta$. 
PARSEC \cite{casale2019parsec} replaces LSTM with a categorical distribution, where $\mu(\theta)$ is a probability vector.
ProxylessNAS \cite{cai2018proxylessnas} and GDAS \cite{xuanyi2019gdas} aim to incorporate the gradient information $\nabla_{\bm{a}_{i}} R(\bm{a},\bm{w})$ rather than a single reward value $R(\bm{a},\bm{w})$ to update $\theta$, as in the case of DP-NAS.
To avoid dense architecture evaluation, they require different approximation techniques (\cite{NIPS2015_5647} and \cite{maddison2017concrete}, respectively) and the resulting parameter update procedures require only sparse architecture evaluation.

\subsubsection{Characteristics of DP-NAS and SP-NAS}
The dense forward or backward propagation is a major drawback of DP-NAS.
The forward propagation results of all the candidate operations at all the edges need to be obtained to compute $\nabla_\theta R(\mu(\theta), \bm{w})$. 
Thus, the GPU memory consumption in each iteration is approximately proportional to $\sum_{i\in\mathcal{I}}\abs{\mathcal{J}_i}$, which prevents the construction of a large parent network.

A popular approach for overcoming this limitation is to optimize the architecture of a smaller network on a proxy task \cite{Zoph2017LTA} consisting of a similar to the original dataset but smaller one. For example, CIFAR \cite{cifar} is used as a proxy task for ImageNet \cite{deng2009imagenet}.
The obtained architecture is expanded by repeating it or increasing the number of nodes to construct a larger network.
Approaches based on proxy tasks rely on transferability.
However, the architectures obtained on the proxy task do not necessarily perform well on other tasks \cite{Dong2020NAS-Bench-201}.
In addition, constructing a proxy task for DP-NAS requires human effort, and the incurred cost prevents practitioners from using NAS in commercial applications.

The advantage of SP-NAS over DP-NAS is that it needs to perform forward (and backward) propagation only in one operation at each edge.
Therefore, its GPU memory consumption in each iteration is independent of the number of operations $|\mathcal{J}_i|$ at each edge.
As its GPU memory consumption is comparable to that of a single architecture, it does not need any proxy task even when a search space designed for a task requires a large parent network.
The advantage of such a search space with a large parent network is demonstrated by, for example, ProxylessNAS \cite{cai2018proxylessnas}, which achieves SOTA accuracy on ImageNet under mobile latency constraints.

\providecommand{\anas}{AdvantageNAS}
\section{\anas}

We propose a novel search strategy for one-shot SP-NAS.
Inspired by the credit assignment problem of policy gradients, we improve the convergence speed of the REINFORCE algorithm, i.e., the baseline search strategy for ENAS, by introducing the so-called advantage function. 
As with other SP-NAS, our proposed approach, \anas, has low GPU memory consumption.

\subsection{Framework}

The probability distribution for $\bm{a}$ is modeled by an independent categorical distribution.
Each $\bm{a}_i$ for $i \in \mathcal{I}$ independently follows a categorical distribution $p_{\theta_i}$ parameterized by $\theta_i$.
The probability mass function for $\bm{a}$ is written as the product $p_{\theta}(\bm{a}) = \prod_{i \in \mathcal{I}} p_{\theta_i}(\bm{a}_i)$, where $\theta = (\theta_1, \dots, \theta_{\abs{\mathcal{I}}})$. The domain of $\theta_i$ is $\R^{\abs{\mathcal{J}_i}}$ and the probability of $\bm{a}_i$ being $1$ in the $j$th category is $\mu_i^{j}(\theta_i)$ defined by \eqref{softmax}. 

The objective function of \anas{} is the reward function expected over the architecture distribution, i.e., $J(\theta, \bm{w}) = \E_{\bm{a} \sim p_\theta}[ R(\bm{a}, \bm{w})]$, similar to REINFORCE. 
It is easy to see that the maximization of $J(\theta, \bm{w})$ leads to the maximization of $R(\bm{a}, \bm{w})$ since
$J(\theta, \bm{w}) \leqslant \max_{\bm{a}} R(\bm{a}, \bm{w})$ and $\sup_{\theta} J(\theta, \bm{w}) = \max_{\bm{a}} R(\bm{a}, \bm{w})$,
where $p_\theta$ represents the Dirac delta function concentrated at $\argmax_{\bm{a}} R(\bm{a}, \bm{w})$.

\anas{} update $\bm{w}$ and $\theta$ alternately. 
The weights $\bm{w}$ is updated by estimating the gradient $\nabla_{\bm{w}}J(\theta, \bm{w})$ using the Monte Carlo method with an architecture sample $\bm{a} \sim p_\theta$ and a minibatch reward $r(\bm{a}, \bm{w}; D) = - \frac{1}{\abs{D}}\sum_{d \in D} \mathcal{L}(\bm{a}, \bm{w}; d)$, where $D \subseteq \mathcal{D}$. Then, $\theta$ is updated by estimating the policy gradient $\nabla_{\theta}J(\theta, \bm{w})$ using the Monte Carlo method with another architecture sample $\bm{a} \sim p_\theta$ and a minibatch reward $r(\bm{a}, \bm{w}; V)$, where $V \subseteq \mathcal{V}$. Next, we discuss how to estimate the policy gradient to improve the convergence speed while maintaining the time and space complexities at levels similar to those in previous studies. 
The algorithm of AdvantageNAS is described in \Cref{pseudo-algo} using pseudocode.

\subsection{Estimation Variance of Policy Gradient}

The REINFORCE algorithm estimates the gradient using a simple Monte Carlo method,
\begin{equation}
  \begin{split}
    \nabla_{\theta} J(\theta, \bm{w})
    &= \E_{\bm{a} \sim p_\theta}[ R(\bm{a}, \bm{w}) \nabla_{\theta} \log(p_\theta(\bm{a}))]
    \\
    &\approx r(\bm{a}, \bm{w}; V) \nabla_{\theta} \log(p_\theta(\bm{a})) \enspace,
  \end{split}
  \label{eq:pg}
\end{equation}
where $\bm{a}$ is an architecture sampled from $p_\theta$.
In general, this gradient is called a \emph{policy gradient}, where $p_\theta$ is regarded as a policy that takes no input (i.e., no state) and outputs the probability of selecting each architecture (as an action).
For our choice of $p_\theta$, the gradient of the log-likelihood is $\nabla_{\theta} \log(p_\theta(\bm{a})) = \bm{a} - \mu(\theta)$.
In other words, the probability vector $\mu(\theta)$ is updated to approach $\bm{a}$ if the minibatch reward $r(\bm{a}, \bm{w}; V)$ is positive. A higher reward results in a greater shift of $\mu(\theta)$ towards current sample $\bm{a}$.

Intuitively, a high estimation variance will lead to slow convergence.
This is indeed theoretically supported by the subsequent result, whose proof follows \cite{akimoto2012algo}; the details are presented in \Cref{apdx:lemma:theta}.
The lower the estimation variance $\Var_{\delta}\left[\delta\right]$ of policy gradient $\delta$, the greater is the lower bound of the expected improvement in one step.
Therefore, reducing the estimation variance is key to speed up the architecture update.
\begin{theorem}\label{lemma:theta}
  Let $J(\theta, \bm{w}) = \E[ R(\bm{a}, \bm{w}) ]$ and $\delta$ be an unbiased estimator of $\nabla_{\theta} J(\theta, \bm{w})$.
  Suppose that $R(\bm{a}, \bm{w}) > 0$ for any $\bm{a}$. 
  Then, for any $\epsilon > 0$,
  \begin{multline}
    \E_{\delta}[ \log J(\theta + \epsilon \cdot \delta, \bm{w} ) ] - \log J(\theta, \bm{w}) \geqslant
    \\
    \Big(\frac{\epsilon}{J(\theta, \bm{w})} - \frac{\epsilon^2}{2} \Big) \norm{\nabla_\theta J(\theta, \bm{w})}^2 - \frac{\epsilon^2}{2} \Var_{\delta}\left[\delta\right]
    \enspace .
  \end{multline}
\end{theorem}

Credit assignment in reinforcement learning tasks is a well-known cause of slow convergence of policy gradient methods \cite{Minsky:1961:ire}.
The reward is the result of a set of consecutive actions (in our case, the choice $\bm{a}_i$ at each edge $i \in \mathcal{I}$).
A positive reward for $\bm{a} = (\bm{a}_1,\dots,\bm{a}_{\abs{I}})$ does not necessarily mean that all the actions $\bm{a}_i$ are promising.
However, since \eqref{eq:pg} updates $\theta$ based on a single reward, it cannot recognize the contribution of each edge.
To understand this problem, consider a situation in which we have two edges with two candidate operations for each edge.
The reward is assumed to be the sum of the contributions of the choices on the two edges, $r_i^{j}$ for $\bm{a}_i = \bm{e}_j$ for $i=1$ and $2$ and $j = 1$ and $2$.
If $r_1^{1} \gg r_2^1 > 0$ and $r_1^{2} \ll r_2^2 < 0$, i.e., the contribution of the first edge dominates, the sign of the reward is determined by $\bm{a}_1$.
The update direction of $\mu_2(\theta_2)$ is determined by $\bm{a}_1$; it is not based on whether $\bm{a}_2$ is a better choice.
This results in a high variance for the $\mu_2(\theta_2)$ update, leading to slow convergence.

\begin{algorithm}[t]
  \caption{AdvantageNAS}
\label{pseudo-algo}
  \begin{algorithmic}[1]
  \While{Until Search Process End}
    \State Sample a minibatch $D$ from $\mathcal{D}$
    \Comment{$\bm{w}$ update step}
    \State Sample an architecture $\bm{a} \sim p_\theta$
    \State Compute a minibatch gradient: \hphantom{XXXXXXXXXXX} \hskip\algorithmicindent \hphantom{XXXX} $\delta_{\bm{w}} \leftarrow \nabla_{\bm{w}} r(\bm{a}, \bm{w}; D)$
    \State Update $\bm{w}$ with some SGD variant using $\delta_{\bm{w}}$
    \State Sample a minibatch $V$ from $\mathcal{V}$
    \Comment{$\theta$ update step}        
    \State Sample an architecture $\bm{a} \sim p_\theta$
    \State Compute minibatch advantages using \Cref{eq:adv-new}: \hskip\algorithmicindent \hphantom{XXXX} $\bar{A}_i \leftarrow A_i(\bm{a}, \bm{w}; V)$ for all $i \in \mathcal{I}$
    \State Compute minibatch gradients: \hphantom{XXXXXXXXXXX} \hskip\algorithmicindent \hphantom{XXXX} $\delta_{\theta_{i}} \leftarrow \bar{A}_i \nabla_{\theta_{i}}\log(p_{\theta}(\bm{a}))$ for all $i \in \mathcal{I}$
    \State Update $\theta_{i}$ with some SGD variant using $\delta_{\theta_{i}}$ ($\forall i \in \mathcal{I}$)
    \EndWhile
  \end{algorithmic}
\end{algorithm}

\subsection{Advantage}\label{sec:adv}

We introduce the notion of \emph{advantage}, used in policy gradient methods, to address the issue of credit assignment \cite{Minsky:1961:ire}.
We regard $\bm{s}_i = (\bm{a}_1,\dots, \bm{a}_{i-1}, \bm{a}_{i+1}, \dots, \bm{a}_{\abs{I}})$ as a state and $\bm{a}_i$ as an action.
By introducing the advantage, we evaluate the goodness of the choice $\bm{a}_i$ at each edge $i$ rather than evaluate the goodness of successive actions $\bm{a}$ by a scalar reward $r$.
We expect to reduce the estimation variance of the policy gradient by the advantage.
  
We define the advantage of $\bm{a}_i$ as (by letting $\bm{a} = (\bm{s}_i, \bm{a}_i)$)
\begin{equation}
  A_i(\bm{a}, \bm{w}; V)
  = r(\bm{a}, \bm{w}; V) - r((\bm{s}_i, \bm{0}), \bm{w}; V)  
  \enspace.\label{eq:adv-e}
\end{equation}
It is understood as follows. Given the state $\bm{s}_i$, the advantage of the action $\bm{a}_i$ is the rewards improvement over $\bm{a} = (\bm{s}_i, \bm{0})$, where the output of the $i$th edge is zero.%
\footnote{When a zero operation (returns zero consistently) is included in the search space, we replace the advantage of the zero operation with the worst expected advantage of the other operations estimated by a moving average. The rationale for this modification is as follows. First, \eqref{eq:adv-e} for the zero operation is always zero, resulting in no architecture update, which is not preferred. Second, the zero operation can mostly be represented by other operations with specific weights. Hence, there is no reason to prefer it unless we consider a latency constraint. See \Cref{app-zero-op} for further details.}
We replace the reward with the advantage in the policy gradient estimation in \eqref{eq:pg}.
If we replace the second term on the right-hand side of \eqref{eq:adv-e} with a single scalar independent of $\bm{s}_i$, the resulting policy gradient is known to be an unbiased estimator \cite{evans2000approximating}. 
However, a state-dependent advantage such as \eqref{eq:adv-e} is not unbiased in general.
The following result states that the policy gradient estimate using this advantage is an unbiased estimate of \eqref{eq:pg}.
Its proof is presented in \Cref{apdx:proof:prop}.
\begin{theorem}%
    \label{thm:unbiasedness}
    Let $\bm{a} \sim p_{\theta}(\bm{a})$ and $d \in V$ be i.i.d.~from $\mathcal{P}$.
    Then, with $A_i$ defined in \eqref{eq:adv-e}, $A_i(\bm{a}, \bm{w}; V) \nabla_{\theta_i} \ln p_{\theta}(\bm{a})$ is an unbiased estimate of $\nabla_{\theta_i}J(\theta, \bm{w})$, i.e., $\nabla_{\theta_i} J(\theta, \bm{w}) = \E_{\bm{a},V} [A_i(\bm{a}, \bm{w}; V) \nabla_{\theta_i} \ln p_{\theta}(\bm{a}) ]$.
  \end{theorem}

  Thus, we can use $\abs{\mathcal{I}}$ different advantage values rather than a single reward value. 
  A shortcoming of the above approach is that it requires the computation of an additional reward to compute $r((\bm{s}_i, \bm{0}), \bm{w}; V)$ for each $i \in\mathcal{I}$; hence, it requires $\abs{\mathcal{I}}$ additional forward propagation to estimate \eqref{eq:pg}.
  As our objective is to reduce the time complexity by improving the convergence speed of SP-NAS, we aim to compute an advantage with ideally no additional computational cost. 
  
  To reduce the computational cost, we approximate the advantage function \eqref{eq:adv-e} as (dropping $\bm{w}$ and $V$ for short)
  \begin{equation}
    \begin{split}
  A_i(\bm{a})
  \approx \nabla_{\bm{a}_i} r(\bm{a}) \rvert_{\bm{a} = (\bm{s}_i, \bm{a}_i)}^\T \bm{a}_{i}
  = \frac{\partial r(\bm{a})}{\partial \mathcal{O}_i^\T} \mathcal{O}_i^j(x_{Q(i)})
    \label{eq:adv-new}
    \enspace,
    \end{split}
\end{equation}
where $\mathcal{O}_i$ and $\mathcal{O}_i^j$ are as in \eqref{forward} and $j \in \mathcal{J}_i$ is the index of the selected operation at edge $i$, i.e., $\bm{a}_i$ is the one-hot vector with $1$ at $j$th.
Also note that here we consider the situation where $\mathcal{O}_i$ and $\mathcal{O}_i^j$ are column vectors and the batch size equals 1 for simplicity. 
The above approximation follows the first-order Taylor approximation of the reward around $\bm{a}$.
Compared to a naive policy gradient computation that uses only a single reward $r(\bm{a}, \bm{w}; V)$,
the policy gradient estimate with the above approximate advantage additionally requires only one reward gradient computation, $\nabla_{\mathcal{O}_i} r(\bm{a}, \bm{w}; V)^\T \mathcal{O}_i^j(x_{Q(i)}; w_{i}^{j})$. The output $\mathcal{O}_i^j(x_{Q(i)}; w_{i}^{j})$ of the selected operation is computed in the forward process and $\nabla_{\mathcal{O}_i} r(\bm{a}, \bm{w}; V)$ is computed in the backward process.
Therefore, our advantage involves no additional computational cost.

\anas\ uses the gradient information of the reward rather than a scalar reward.
Existing works on SP-NAS also use the gradient \cite{cai2018proxylessnas,xuanyi2019gdas}.
\anas\ aims to reduce the estimation variance while keeping the policy gradient unbiased,
whereas the previous methods simply try to incorporate the gradient information of the reward and they do not consider the estimation bias.

\subsection{Variance Reduction Effect : An Example}\label{sec:vr}

We know from \Cref{thm:unbiasedness} that the advantage \eqref{eq:adv-e} provides an unbiased estimator of the gradient $\nabla_{\theta} J(\theta, \bm{w})$ and from \Cref{lemma:theta} that reducing the variance in an unbiased estimator of the gradient contributes toward improving the lower bound of the expected improvement.
Since \eqref{eq:adv-new} is an approximation of \eqref{eq:adv-e}, we expect that our advantage contributes toward reducing the estimation variance, resulting in faster convergence.

To understand the variance reduction effect of the proposed advantage, we consider a simple situation in which the reward is given by $\bm{r}^\T \bm{a} = \sum_{i=1}^{\abs{\mathcal{I}}} \bm{r}_i^\T \bm{a}_i$ for some vector $\bm{r} = (\bm{r}_1, \dots, \bm{r}_{\mathcal{I}})$.
In this example, neither $\mathcal{V}$ nor $\bm{w}$ is introduced.
In this linear objective situation, the advantages \eqref{eq:adv-e} and \eqref{eq:adv-new} are equivalent and $A_i(\bm{a}) = \bm{r}_i^\T \bm{a}_i$. The square norm of the gradient component $\nabla_{\theta_i} J(\theta)$ is $\sum_{j\in \mathcal{J}_i} (r_{i}^{j})^2 (\mu_i^j)^2 (1 - \mu_i^j)^2$.
The estimation variance of the policy gradient $\delta_{\mathrm{adv}}$ using the advantage \eqref{eq:adv-e} or \eqref{eq:adv-new} is $\Var[\delta_{\mathrm{adv}}] = \sum_j \mu_{i}^{j} (r_{i}^j)^2 (1 - 2\mu_{i}^{j} + \norm{\mu_i}^2)$.
Meanwhile, the estimation variance of the policy gradient $\delta_{\mathrm{base}}$ using a reward value (i.e., \eqref{eq:pg}) is $\Var[\delta_{\mathrm{base}}] = \Var[\delta_{\mathrm{adv}}] + ( 1 - \norm{\mu_i}^2) \sum_{k\neq i} \sum{j} (r_k^j)^2 \mu_k^j$.
In other words, the variance reduction effect for each gradient component typically scales linearly with $\abs{\mathcal{I}}$ (because $\sum_{k\neq i}$), whereas $\norm{\nabla_{\theta_i} J(\theta)}^2$ and $\Var[\delta_{\mathrm{adv}}]$ do not.
In light of \Cref{lemma:theta}, the proposed advantage potentially contributes toward improving the search performance, especially when the number of edges is $\abs{\mathcal{I}} \gg 1$.

\section{Experiments}

We performed three experiments to compare the search efficiency of \anas{} with that of existing gradient-based search strategies for NAS.
Note that our objective is not to obtain the SOTA architecture; hence, the search spaces used in the experiments were relatively small for convenience.
To compare the search strategies for NAS fairly, we implemented all the baseline NAS methods on the same code base and imposed the same setting except for the search strategy.

\subsection{Toy Regression Task}\label{sec:toy}
\subsubsection{Motivation}
Whereas one-shot NAS methods optimize the weights and architecture simultaneously, to solely assess the performance improvement due to the differences in the architecture updates, we evaluate the performance of the architecture optimization while the weights are fixed.
In particular, we confirm that the variance reduction by our advantage contributes toward accelerating the architecture search by comparing the performance of REINFORCE and \anas, which are different only in terms of the reward computation for each edge as described in \Cref{sec:adv}.

\subsubsection{Task Description}
We prepare two networks: teacher-net $\phi_{\bm{a}^*}$ and student-net $\phi_{\bm{a}}$.
Both networks are represented by the same DAG.
The two networks have $|\mathcal{I}| = 10$ intermidiate nodes, and there are edges from one input node to all the intermidiate nodes.
For each edge $i \in \mathcal{I}$, $\abs{\mathcal{J}_i} = 10$ different convolutional filters are prepared as the candidate operations.
These filters have a filter size of $7\times7$, stride of $2$, and output channels of $1$ followed by $\tanh$ activation.
The architecture variable $\bm{a}^*$ of teacher-net is such that $\bm{a}_i^*$ are one-hot vectors, i.e., one convolutional filter per edge which connects the input node and the intermediate node.
For each edge from an intermediate node to the output node, a convolution with filter size of $4\times4$, stride of $1$, and output channels of $1$ is applied.
The average of the outputs is taken as the aggregation function for the output node.
A minibatch $V$ with batch size of $100$ is sampled randomly for each iteration, where the input data $\bm{x}$ is generated by the uniform distribution over $[-1,+1]^{13 \times 13 \times 1}$ and its ground truth is given by $\phi_{\bm{a}^*}(\bm{x})$.
The reward function is defined as the negative loss; the squared error between the ground truth $\phi_{\bm{a}^*}(\bm{x})$ and the student prediction $\phi_{\bm{a}}(\bm{x})$. 
The weights for the convolutions are fixed to be the same ($\bm{w}^*$) for both networks during the architecture optimization process.
The $\bm{w}^{*}$ and $\bm{a}^{*}$ are defined randomly before the optimization process.
The loss is consistently zero if and only if student-net learns the optimal architecture $\bm{a}^*$.
We optimize $\theta$ using the Adam optimizer \cite{kingma2014adam} with a learning rate of $0.001$ and $\beta=(0.9, 0.999)$.
The implementation details of gradient-based methods are presented in \Cref{sec:grad-impl}.

\begin{figure}[t]
  \centering
 \includegraphics{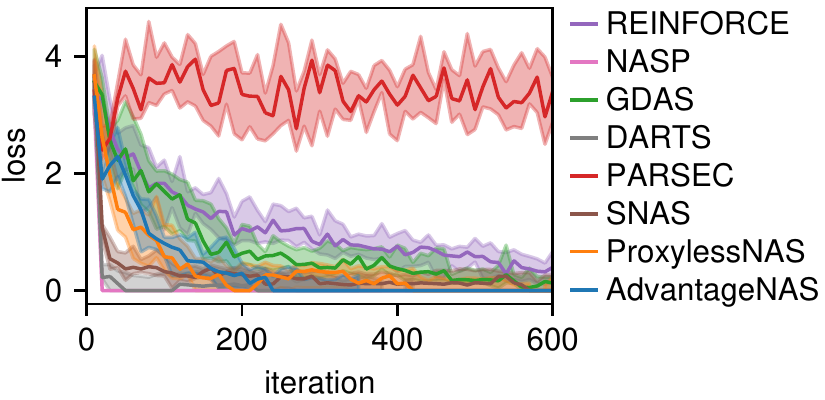}
  \caption{Results of the toy regression task. Median (solid line) and 25th—75th percentile area (colored area) over 10 trials.}
  \label{fig-toy1}
\end{figure}

\subsubsection{Results}
\Cref{fig-toy1} shows the results.
The test loss is computed for the architecture obtained by selecting the operation at each edge such that $\mu_i(\theta_i)$ is maximum.
We observe that \anas{} converges faster than existing SP-NAS such as REINFORCE, PARSEC and GDAS.
Moreover, it is more stable than ProxylessNAS when the number of iterations is large.
Existing DP-NAS approaches such as DARTS, SNAS and NASP converge faster than \anas.
However, as mentioned above, these methods require nearly $|\mathcal{J}|$ times the memory space to perform the architecture search, which prevents their application to a large-scale task requiring a large architecture.

\providecommand{\acc}{accuracy}
\providecommand{\param}{params(MB)}

\begin{table*}[h]
\centering
\small
      \begin{tabular}{l|c|c|c|c}
        \hline

        \hline
        
        \multirow{2}{*}{Method} & \multicolumn{2}{c}{\texttt{argmax}} & \multicolumn{2}{|c}{\texttt{process best}} \\
        \cline{2-5}
                                & \acc & \param & \acc & \param \\
                         \hline
        
        \hline
        REINFORCE\cite{pham2018enas,Williams1992} & $ 93.01 \pm 0.47 $ & $ 0.67 \pm 0.18 $ & $ 92.69 \pm 0.52 $ & $ 0.68 \pm 0.20 $ \\
        PARSEC\cite{casale2019parsec} & $ 93.68 \pm 0.12 $ & $ 1.39 \pm 0.20 $ & $ 92.51 \pm 0.65 $ & $ 0.47 \pm 0.12 $ \\
        ProxylessNAS\cite{cai2018proxylessnas} & $ 93.22 \pm 0.08 $ & $ 0.74 \pm 0.10 $ & - & - \\
        GDAS\cite{xuanyi2019gdas} & $ 93.46 \pm 0.13 $ & $ 1.24 \pm 0.10 $ & $ 92.53 \pm 0.98 $ & $ 0.62 \pm 0.20 $ \\
        \hline
        AdvantageNAS & $ {\bf 93.76 \pm 0.00} $ & $ 1.53 \pm 0.00 $ & $ {\bf 93.29 \pm 0.21} $ & $ 0.94 \pm 0.11 $ \\
        \hline

        \hline
      \end{tabular}
\caption{Results on CIFAR-10. Average$\pm$std accuracy over three search trials as well as the parameter size associated with the final architecture.
Note that \texttt{process best} for ProxylessNAS is missing as it samples two operations at each edge and the one-hot architecture is never evaluated during the architecture search process, as described in \Cref{tab:related}.}\label{tab-cifar}
\end{table*}

\subsection{Convolutional Cell on CIFAR-10} \label{ex:cifar}

\begin{figure}[t]
  \centering
  \includegraphics{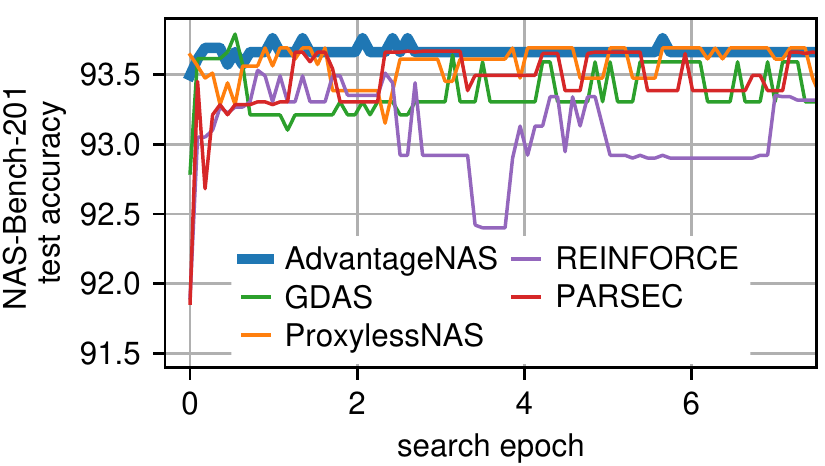}
  \caption{Improvement progress of search process. All the lines represent the median for three search trials. The values are obtained by querying NAS-Bench-201.}\label{cifar-10-main}
\end{figure}

\subsubsection{Motivation}
We compare AdvantageNAS with existing SP-NAS by searching for a convolutional cell on the CIFAR-10 \cite{cifar} dataset and evaluating them on the NAS-Bench-201 \cite{Dong2020NAS-Bench-201}.

\subsubsection{Search Settings}
The CIFAR-10 dataset contains 50000 training data and 10000 test data.
We split the original training data into 25000 training data $\mathcal{D}$ and 25000 validation data $\mathcal{V}$, as in the case of NAS-Bench-201.
We use $\mathcal{D}$ to optimize $\bm{w}$ and $\mathcal{V}$ to optimize $\theta$.
Our architecture search space is the same as NAS-Bench-201 except that the zero operations are removed.\footnote{Our preliminary experiments showed that zero operations disrupt the architecture search of existing SP-NAS (e.g., ProxylessNAS ended up with a final accuracy of $10\%$). We removed zero operations to avoid the effects of search space definition. See \Cref{apdx:zero-result} for the results on the search space including the zero operations.}
We search for the convolutional cell architecture and stack it 15 times to construct a macro skeleton.
All the stacked cells have same architecture.
The cell structure has a total of $|\mathcal{I}| = 6$ edges.
In each cell, there are $|\mathcal{J}| = 5$ candidate operations such as $1\times1$ convolution, $3\times3$ convolution, $3\times3$ average pooling and identity map.
The total procedure was conducted in 100 epochs; however, $\theta$ was fixed in the first 50 epochs for pre-training $\bm{w}$.
The other settings are presented in \Cref{app-cifar}.

\begin{table*}[t]
\centering
\small
\begin{tabular}{l|c|c|c}
  \hline

  \hline
Method & validation perplexity & test perplexity & \param \\
  \hline

  \hline
REINFORCE\cite{pham2018enas,Williams1992} & $99.48 \pm 1.51$ & $95.50 \pm 1.73$ & $7.40 \pm 0.07$ \\
PARSEC\cite{casale2019parsec} & $99.68 \pm 1.49$ &  $95.51 \pm 1.01$ & $7.33 \pm 0.09$  \\
ProxylessNAS\cite{cai2018proxylessnas} & $95.12 \pm 0.22$ & ${\bf 90.76 \pm 0.23}$ & $7.33 \pm 0.09$  \\
GDAS\cite{xuanyi2019gdas} & $99.28 \pm 2.00$ & $95.40 \pm 2.01$ & $7.33 \pm 0.09$ \\
\hline
\anas & ${\bf 94.90 \pm 0.18}$ & $91.18 \pm 0.13$ & $7.44 \pm 0.00$ \\
  \hline

  \hline
\end{tabular}

\caption{Results on PTB. Average$\pm$std accuracy over five search trials as well as the parameter size associated with the final architecture.}\label{ptb-tab}
\end{table*}

\subsubsection{Evaluation Settings}
To evaluate the performance of one-shot NAS, the weights for the architectures obtained by the search process are often retrained for longer epochs; then, the test performances are compared.
To minimize the influence of the quality of retraining on the performance metric, we use NAS-Bench-201 \cite{Dong2020NAS-Bench-201} for evaluation.
It is a database containing average accuracy for all architectures in the search space.
The accuracy was computed after 200 epochs of training on CIFAR-10 with each architecture.
Each architecture was trained three times independently and its average accuracy is recorded in the database. 
We use this average accuracy for the performance metric.

There are different ways for determining the final architecture after architecture search.
One way is to choose the best architecture based on its validation loss observed in the search process. This way is adopted by, e.g., ENAS (REINFORCE) and GDAS, and is denoted as \texttt{process best}. 
Another way (adopted by ProxylessNAS and PARSEC) is to take operation $j$ for which $\mu_i^{j}(\theta_i)$ is maximum at each edge $i$, as in \Cref{sec:toy}.
It is denoted as \texttt{argmax}.
In our experiment, we show the performance of these two ways and confirmed which one is better suited for \anas.

\subsubsection{Results}
\Cref{tab-cifar} presents the results.
\anas{} achieves an improvement over REINFORCE solely owing to the introduction of the advantage described in \Cref{sec:adv}.
Moreover, \anas{} outperforms all the existing SP-NAS in terms of both \texttt{argmax} and \texttt{process best}.
Although \texttt{process best} was adopted in ENAS and GDAS, \texttt{argmax} delivered better performances in all the algorithms including \anas.
A comparison with DP-NAS is shown in \Cref{apdx:dp-nas} for reference.

The search progress in the initial steps is shown in \Cref{cifar-10-main}.
The median of the average accuracy on NAS-Bench-201 is shown for each method with \texttt{argmax} over three independent search trials.
At the early stage in the search process, AdvantageNAS achieved a higher accuracy than that of the existing SP-NAS and was more stable at the high accuracy.
It can be inferred that AdvantageNAS outperforms the existing SP-NAS in terms of convergence speed.

To further assess the convergence speed, we analyzed the entropy of categorical distribution $\mu(\theta)$ in the search process.
The details are presented in \Cref{apdx:entropy}.

\subsection{Recurrent Cell on Penn Treebank}
\subsubsection{Motivation} We confirm that the usefulness of \anas{} for recurrent cell architecture search as well as for convolutional cells.

\subsubsection{Settings}
We use Penn Treebank \cite{marcus1993ptb} (PTB) dataset to search for the recurrent cell architecture.
Most of the settings is the same as those of DARTS \cite{liu2018darts}. See \Cref{app-ptb} for the details.
The recurrent cell structure has $|\mathcal{I}| = 6$ edges.
They have $|\mathcal{J}| = 5$ operation candidates, namely ReLU, tanh, sigmoid, identity, and zero operation, after linear transformation is applied.
The parameters of the linear transformation are trainable weights.
In the search process, we use the training data of PTB to optimize both the weights and $\theta$ for 30 epochs. 
Finally, the obtained architectures are retrained in 100 epochs.
According to previous studies, we chose the final architecture by \texttt{argmax} in ProxylessNAS and PARSEC and by \texttt{process best} for REINFORCE(ENAS) and GDAS.
The final architecture of AdvantageNAS is determined by \texttt{argmax}.
The validation and test data of PTB are used to measure the final performance.

\subsubsection{Results}
The performances of the retrained architectures are shown in \Cref{ptb-tab}.
AdvantageNAS achieved highest performance in validation data.
In addition, AdvantageNAS achieved the competitive result obtained by ProxylessNAS in test data.
This result implies that AdvantageNAS can be applied to various domains.
The architectures obtained by AdvantageNAS and ProxylessNAS were similar, as shown in \Cref{vis-arc}.

\section{Conclusion}
We proposed AdvantageNAS, a novel gradient-based search strategy for one-shot NAS.
AdvantageNAS aims to accelerate the architecture search process while maintaining the GPU memory consumption as in the case of sparse propagation NAS.
For this purpose, we introduced the notion of \emph{credit assignment} into the architecture search to reduce the variance of gradient estimation.
A unique aspect of our study is the theoretical guarantee of the monotonic improvement of the expected loss by AdvantageNAS.
Empirical evaluations were performed on different datasets and search spaces, confirming the promising performance of AdvantageNAS.
To compare only the search strategies, we implemented the existing approaches on the same publicly available code base.
To make an important contribution to the NAS community, we focused on investigating only the efficacy of the search strategy. Finding SOTA architectures, which requires searching for best combinations of different NAS components, is beyond the scope of this study.
Further investigation of larger search spaces is an important direction for future research.

\section*{Acknowledgements}

The first author would like to thank Idein Inc. for their support.
This paper is partly based on the results obtained from a project commissioned by the New Energy and Industrial Technology Development Organization (NEDO).

\bibliography{nas}

\clearpage
\appendix

\section{Proofs}\label{apdx:proof}

\subsection{Proof of \Cref{lemma:theta}}\label{apdx:lemma:theta}

  Let $D_\mathrm{KL}(q \parallel q') = \int q(x) \log (q(x) / q'(x)) dx$ be the KL-divergence between $q$ and $q'$.
  Let
  \begin{align*}
    q_\theta(\bm{a}; \bm{w}) &= R(\bm{a}, \bm{w}) p_{\theta}(\bm{a}) / J(\theta, \bm{w})\enspace, \\
    q_{\theta'}(\bm{a}; \bm{w}) &= R(\bm{a}, \bm{w}) p_{\theta'}(\bm{a}) / J(\theta', \bm{w})  \enspace.
  \end{align*}
  It has been shown in \cite{akimoto2012algo} that
  \begin{equation}
    \begin{split}
    \MoveEqLeft[1]\log J(\theta', \bm{w} ) - \log J( \theta, \bm{w}) \\
    &= D_\mathrm{KL}( q_\theta(\bm{a}; \bm{w}) \parallel q_{\theta'}(\bm{a}; \bm{w}))\\
    &+ \int q_\theta(\bm{a}; \bm{w}) \log p_{\theta'}(\bm{a}) d\bm{a} \\
    &- \int q_\theta(\bm{a}; \bm{w}) \log p_{\theta}(\bm{a}) d\bm{a} \enspace,
    \end{split}
  \end{equation}
  where the first term is the KL-divergence and is guaranteed to be non-negative.
  Letting $\theta' = \theta + \epsilon \cdot \delta$, the second and third terms reduce to %
  \begin{align*}
    \MoveEqLeft[1]
    \int q_\theta(\bm{a}; \bm{w}) \log p_{\theta'}(\bm{a}) d\bm{a}
    - \int q_\theta(\bm{a}; \bm{w}) \log p_{\theta}(\bm{a}) d\bm{a}
    \\
 &= \int_{0}^{\epsilon} \frac{\partial}{\partial \tau}\left(\int q_\theta(\bm{a}; \bm{w}) \log p_{\theta + \tau \cdot \delta}(\bm{a}) d\bm{a}\right) d\tau
    \\
 &= \delta^\T \int_{0}^{\epsilon} \left(\nabla_{\tilde \theta}\int q_\theta(\bm{a}; \bm{w}) \log p_{\tilde \theta}(\bm{a}) d\bm{a}\right)_{\tilde\theta = \theta + \tau \cdot \delta} d\tau \enspace.
  \end{align*}
  The integrand is
  \begin{align*}
    \MoveEqLeft[0]
    \nabla_{\tilde \theta} \int q_\theta(\bm{a}; \bm{w}) \log p_{\tilde \theta}(\bm{a}) d\bm{a} 
    \\
 &= 
   \int q_\theta(\bm{a}; \bm{w}) \nabla_{\tilde \theta} \log p_{\tilde \theta}(\bm{a}) d\bm{a}
    \\
 &= 
   \int q_\theta(\bm{a}; \bm{w}) \left(\bm{a} - \E_{\tilde\theta}[\bm{a}] \right) d\bm{a}
    \\
 &= 
   \frac{1}{ J(\theta, \bm{w})}\int R(\bm{a}, \bm{w}) p_{\theta}(\bm{a}) \left(\bm{a} - \E_{\theta}[\bm{a}] \right) d\bm{a}
    \\
 &\quad 
   + \frac{1}{ J(\theta, \bm{w})}\int R(\bm{a}, \bm{w}) p_{\theta}(\bm{a}) \left(\E_{\theta}[\bm{a}] - \E_{\tilde\theta}[\bm{a}] \right) d\bm{a}
    \\
 &= 
   \frac{ \nabla_\theta J( \theta, \bm{w}) }{ J( \theta, \bm{w})}
   + \left(\E_{\theta}[\bm{a}] - \E_{\tilde\theta}[\bm{a}] \right) \enspace.
  \end{align*}
  Hence,
  \begin{align*}
    \MoveEqLeft[2]
    \int q_\theta(\bm{a}; \bm{w}) \log p_{\theta'}(\bm{a}) d\bm{a}
    - \int q_\theta(\bm{a}; \bm{w}) \log p_{\theta}(\bm{a}) d\bm{a}
    \\
 =& \frac{\epsilon}{J(\theta, \bm{w})} \inner{\nabla_\theta J( \theta, \bm{w})}{\delta} \\
 &+ \int_{0}^{\epsilon}
   \inner{\E_{\theta}[\bm{a}] - \E_{\theta + \tau \cdot \delta}[\bm{a}] }{ \delta }
   d\tau \enspace.
  \end{align*}
  The mean value theorem in the integral form leads to
  \begin{align*}
    \MoveEqLeft[1]
   \inner{\E_{\theta}[\bm{a}] - \E_{\theta + \tau \cdot \delta}[\bm{a}] }{ \delta }
    \\
    &= 
   \inner{ \nabla_{\theta_i} \log Z_i(\theta_i) - \nabla_{\tilde\theta_i} \log Z_i(\tilde\theta_i)\mid_{\tilde\theta = \theta + \tau \cdot \delta} }{ \delta }
    \\
    &= -\int_{0}^{\tau}
    \delta^\T      
   \left( \nabla_{\tilde\theta_i}\nabla_{\tilde\theta_i} \log Z_i(\tilde\theta_i) \right)_{\tilde\theta = \theta + s \cdot \delta } 
    \delta ds
    \\
    &=
    -\delta^\T      
   \left( \int_{0}^{\tau} \Cov_{\theta + s \cdot \delta}[\bm{a}_i] ds \right)
    \delta
    \\
    &\geqslant
    -\tau \norm{\delta}^2 \enspace.
  \end{align*}
  We have
  \begin{multline*}
    \int q_\theta(\bm{a}; \bm{w}) \log p_{\theta'}(\bm{a}) d\bm{a}
    - \int q_\theta(\bm{a}; \bm{w}) \log p_{\theta}(\bm{a}) d\bm{a}
    \\
 \geqslant \frac{\epsilon}{J(\theta, \bm{w})}\inner{\nabla_\theta J(\theta, \bm{w})}{\delta} - \frac{\epsilon^2}{2} \norm{\delta}^2 \enspace.
  \end{multline*}
  Taking the expectation on the right-hand side, we obtain
  \begin{multline*}
  \E_\delta\left[\frac{\epsilon}{J(\theta, \bm{w})}\inner{\nabla_\theta J(\theta, \bm{w})}{\delta} - \frac{\epsilon^2}{2} \norm{\delta}^2\right] 
  \\
  = \frac{\epsilon}{J(\theta, \bm{w})} \norm{\nabla_\theta J(\theta, \bm{w})}^2 - \frac{\epsilon^2}{2} \E_{\delta}\left[\norm{\delta}^2\right] 
  \enspace. 
\end{multline*}
This completes the proof. \qed

\subsection{Proof of \Cref{thm:unbiasedness}}\label{apdx:proof:prop}

Since $r(\bm{a}, \bm{w}; V)$ is an unbiased estimate of $R(\bm{a}, \bm{w})$, it suffices to show that
\begin{multline*}
  \E_{\bm{a},V} [ r(\bm{a}_{[\bm{a}_i \to \bm{b}_i]}, \bm{w}; V) \nabla_{\theta_i} \ln p_{\theta}(\bm{a}) ]
  \\= \E_{\bm{a}} [ R(\bm{a}_{[\bm{a}_i \to \bm{b}_i]}, \bm{w}) \nabla_{\theta_i} \ln p_{\theta}(\bm{a}) ] = \bm{0}\enspace.
\end{multline*} 
Note that $\nabla_{\theta_i} \ln p_{\theta}(\bm{a}) = \bm{a}_i - \mu_i(\theta_i)$. 
  Since $\bm{a}_i$ and $\bm{a}_{[\bm{a}_i \to \bm{b}_i]}$ are independent, the expectation of the product of $R(\bm{a}_{[\bm{a}_i \to \bm{b}_i]}, \bm{w})$ and $\bm{a}_i - \mu_i(\theta_i)$ is the product of the expectations $\E_{\bm{a}_{-i}} [ R(\bm{a}_{[\bm{a}_i \to \bm{b}_i]}, \bm{w}) ]$ and $\E_{\bm{a}_i}[ \bm{a}_i - \mu_i(\theta_i) ] = \bm{0}$.
  This completes the proof.  \qed

\section{Advantage for Zero Operation} \label{app-zero-op}

 The zero operation, which outputs a zero tensor consistently, is often included in the candidate operations to represent sparse architectures.
 As the advantage for the zero operation is consistently $0$ by definition, no update $\theta$ occurs when the zero operation is selected in the architecture search step.

 Typically, the zero operation can be represented by non-zero operations by optimizing their weight parameters.
 Therefore, there is no need to prefer zero operation to other operations and include it in search space.
 However, latency constraint techniques \cite{cai2018proxylessnas, xie2018snas} constitute an important research direction for NAS.
Therefore, in the future, to introduce latency regularization into AdvantageNAS, we will provide the modified advantage value for the zero operation as follows.

 The advantage for the zero operation is computed using the exponential moving average of advantages.
 Let $EMA \in \R^{\sum_{i\in\mathcal{I}} \abs{\mathcal{J}_i}}$ be the vector of exponential moving averages, initialized as $\bm{0}$, and let $\gamma > 0$ be the decay factor.
 In each iteration, we update the moving average as
 \begin{equation}
   EMA(i) \leftarrow EMA(i) + \gamma (A_i(\bm{a}, \bm{w}; V)\bm{a}_i - EMA(i) \odot \bm{a}_i)  \enspace, \label{zero-ema}
 \end{equation}
 for all $i \in \mathcal{I}$, where $EMA(i) \in \mathbb{R}^{|\mathcal{J}_i|}$ is a part of $EMA$ corresponding to edge $i$, and $\odot$ denotes the Hadamard product.
 Suppose that $j = 1$ corresponds to the zero operation at edge $i$ and it is selected.
 Then, its advantage is replaced with $\min_{j \in \mathcal{J}_i\setminus\{1\}} EMA(i, j)$, where $EMA(i,j)$ is the element of $EMA$ corresponding to the $j$th operation at edge $i$.
 In other words, we replace the advantage of the zero operations with the worst expected advantage at each edge.
 If no regularization is introduced, the zero operation will not be selected as the final architecture.

 \section{Implementation Details}\label{sec:grad-impl}

For fair comparison of the related gradient methods, we re-implemented all the related methods.
We used the original implementations of GDAS\footnote{\url{https://github.com/D-X-Y/AutoDL-Projects/blob/master/lib/models/cell_searchs/search_cells.py}, \url{https://github.com/D-X-Y/AutoDL-Projects/blob/master/lib/models/cell_searchs/search_model_gdas.py}}, ProxylessNAS\footnote{\url{https://github.com/mit-han-lab/proxylessnas/blob/master/search/modules/mix_op.py}}, and DARTS\footnote{\url{https://github.com/quark0/darts/blob/master/rnn/model_search.py}, \url{https://github.com/quark0/darts/blob/master/rnn/model.py}} as references.

According to the default settings in each method, we set the initial temperature of the Gumbel softmax in GDAS to $10$ and linearly reduced it to $0.1$.
We set the initial temperature of the Gumbel softmax in SNAS to $1$ and reduced it to $0$ with the cosine schedule.
We set the number of Monte Carlo samples in PARSEC to $8$.
We applied the baseline term for REINFORCE. The reward in REINFORCE is modified as $R(\bm{a},\bm{w}) - b$, where $b$ is $\mathbb{E}_{\bm{a}\sim p_{\theta}(\bm{a})}[R(\bm{a},\bm{w})]$, but approximated by the exponential moving average as in \Cref{zero-ema}.
We set the decay factor $\gamma$ of the exponential moving average in \Cref{zero-ema} and the baseline term of REINFORCE to $0.05$.
Our implementation of the gradient methods is common to all our experiments.

\section{Settings for Convolutional Architecture Search}\label{app-cifar}

The macro skeleton design follows the NAS-Bench-201\cite{Dong2020NAS-Bench-201}.
The main body is composed of three stages.
Each stage has five cells; hence, the main body has 15 cells in total, and they have the same architecture.
A basic residual block was inserted between successive stages, which reduces the resolution by half and expands the output channels.
The output channels were set to 16, 32, and 64 for the first, second, and third stages, respectively.
The main body has a convolutional layer as a preprocessing layer, and a global average pooling layer followed by a fully connected layer as a post-processing layer.
The cell has one input node indexed as $0$, one output node indexed as $3$, and two intermediate nodes indexed as $1$ and $2$.
There are edges from node $i$ to $j$ for all $i < j$.
Hence, this cell structure has $6$ edges in total.

We initialize $\theta$ by ones, i.e., equal probability for all candidate operations.
Furthermore, we applied a gradient clipping of $1.0$ to both $\bm{w}$ and $\theta$ for optimization stability.

Following previous studies \cite{liu2018darts,Dong2020NAS-Bench-201}, we configured the search strategy as follows.
$\theta$ and the weights are optimized alternately. 
According to \cite{Dong2020NAS-Bench-201}, the weights were optimized using Nesterov SGD with a momentum of $0.9$, weight decay rate of $5\times10^{-4}$, and learning rate initialized to $0.1$ and annealed down to 0 with a cosine schedule.
According to \cite{liu2018darts}, we optimized $\theta$ using Adam with $\beta=(0.5,0.999)$, weight decay rate of $3\times10^{-3}$, and learning rate initialized to $3\times10^{-4}$ and annealed down to 0 with a cosine schedule.
The batch size was 64 for both $D$ and $V$.
We applied random horizontal flip with a probability of $0.5$
and randomly cropped 32$\times$32 pixels from an image of size 40$\times$40 after 4-pixel padding on each side of the original image of size 32$\times$32.

\section{Settings for Reccurent Architecture Search} \label{app-ptb}

Most of the settings are the same as those in \cite{liu2018darts}.
Our macro skeleton is constructed in three steps.
An input vector of each timestep is converted by a word embedding layer and fed into a recurrent cell.
Finally, the output of the recurrent cell is transformed by a fully connected layer, which is the final output of the macro skeleton.
The recurrent cell has one input node indexed as $0$, $N$ intermediate nodes indexed as $1,2,...,N$, and one output node indexed as $N+1$.
There are edges between node indices $i$ to $j$ where $i<j<N+1$.
We set the number $N$ of intermediate nodes to $3$.
The feature $\bm{x}_{0}$ of the input node is calculated as $\bm{x}_{0} = \bm{h} \oplus \bm{c}_{in} \odot (\tanh(W_{in}^{h}(\bm{x}_{e},\bm{h})) \ominus \bm{h})$, where $\bm{c}_{in} = \mathrm{sigmoid}(W_{in}^{c}(\bm{x}_{e},\bm{h}))$, $\bm{h}$ is the output of the recurrent cell at the previous timestep, $\bm{x}_{e}$ is the output feature of the embedding layer, $\oplus$ and $\ominus$ denote the element-wise plus and minus, respectively, and $\odot$ denotes the Hadamard product.
The feature $\bm{x}_{N+1}$ of the output node is calculated as the element-wise average of the features of all the intermediate nodes.
Further, note that $W_{i}^{c}$ and $W_{i}^{h}$ for any $i$ are the learnable weights.
For each edge, we calculated the output $\bm{y}_{i}$ of edge $i$ as $\bm{y}_{i} =  \bm{x}_{Q(i)} \oplus \bm{c}_{i} \odot (\mathrm{act}(W_{i}^{h}\bm{x}_{Q(i)}) \ominus \bm{x}_{Q(i)})$, where $\bm{c}_{i} = \mathrm{sigmoid}(W_{i}^{c}\bm{x}_{Q(i)})$, $\bm{x}_{Q(i)}$ is the input feature of edge $i$, and $\mathrm{act}$ is an operation selected at edge $i$.
We set the feature size of each nodes to $300$.

We used SGD with an initial learning rate of $20$, which is annealed down to 0 with a cosine schedule for the weight optimization, with a batch size of $256$, gradient clipping of $0.25$, and BPTT length of $35$.
We used the Adam optimizer to update $\theta$ with the learning rate initialized to $3.0\times10^{-3}$ and annealed down to $0$ with a cosine schedule, and $\beta=(0.9,0.999)$. 
We applied dropout for embeddings and recurrent cell outputs with a probability of $0.5$.
Furthermore, we applied gradient clipping of $1$ to the architecture optimization.
The optimization strategy in the retraining phase for evaluation was the same as that in the above search settings.

\begin{table*}[t]
  \centering
  \small
  \begin{tabular}{l|c|c|c|c}
    \hline

    \hline
    \multirow{2}{*}{Method} & \multicolumn{2}{c}{\texttt{argmax}} & \multicolumn{2}{|c}{\texttt{process best}} \\
    \cline{2-5}
                            & \acc & \param & \acc & \param \\
    \hline

    \hline
    REINFORCE\cite{pham2018enas,Williams1992} & $ 65.15 \pm 39.00 $ & $ 0.54 \pm 0.08 $ & $ 91.40 \pm 0.45 $ & $ 0.49 \pm 0.10 $ \\
    PARSEC\cite{casale2019parsec} & $ 92.67 \pm 0.55 $ & $ 0.59 \pm 0.00 $ & $ 91.75 \pm 0.57 $ & $ 0.48 \pm 0.17 $ \\
    ProxylessNAS\cite{cai2018proxylessnas} & $ 10.00 \pm 0.00 $ & $ 0.07 \pm 0.00 $ & - & - \\
    GDAS\cite{xuanyi2019gdas} & $ 88.62 \pm 0.00 $ & $ 0.34 \pm 0.00 $ & $ 92.82 \pm 0.71 $ & $ 0.47 \pm 0.14 $ \\
    \hline
    \anas & $ {\bf 93.63 \pm 0.18} $ & $ 1.46 \pm 0.10 $ & $ {\bf 93.18 \pm 0.19} $ & $ 0.86 \pm 0.20 $ \\
    \hline

    \hline
  \end{tabular}
  \caption{Results on CIFAR-10 including zero operation. See \Cref{tab-cifar} for details.}\label{tab-cifar-apdx}
\end{table*}

\section{SP-NAS Result of Convolutional Architecture Search with Zero Operation}\label{apdx:zero-result}

\Cref{tab-cifar-apdx} shows the result of SP-NAS for convolutional architecture search.
The experimental setting is the same as that for \Cref{tab-cifar} except that we included zero operation in the search space.
Comparison of \Cref{tab-cifar-apdx} and \Cref{tab-cifar} shows how much each approach was affected by zero operation.
The difference was minimal for AdvantageNAS owing to the zero operation treatment described in \Cref{app-zero-op}.
We observed a significant deterioration for the other SP-NAS methods, especially when \texttt{argmax} was used.
As \texttt{argmax} is based on the distribution parameter $\theta$ for the architecture vector after the search process,
it implies that they converged toward rather sub-optimal solutions, which are much worse than the best solutions obtained during the search process (\texttt{process best}).
A comparison of the ``param'' columns of \Cref{tab-cifar-apdx} and \Cref{tab-cifar} shows that SP-NAS approaches except \anas\ tended to architectures having much fewer trainable parameters when zero operations were included.

\section{Result of Convolutional Architecture Search with DP-NAS}\label{apdx:dp-nas}
\begin{figure}[h]
  \centering
  \includegraphics{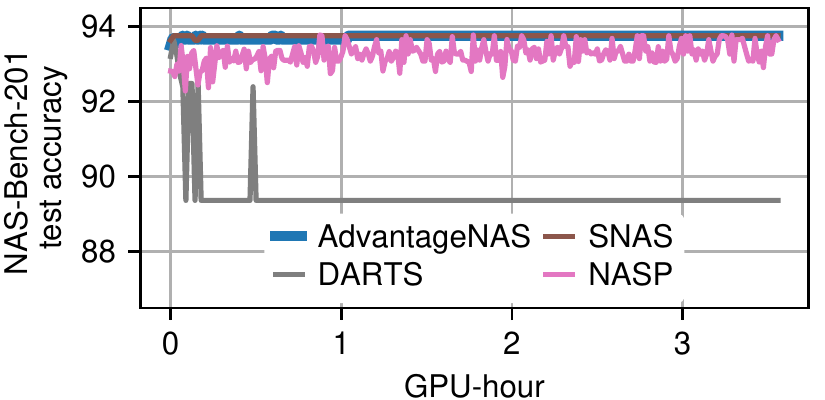}
  \caption{Improvement in progress of search processes. The lines represent the medians over three search trials. The values were obtained by querying NAS-Bench-201.}
  \label{fig:dp-cifar}
\end{figure}

\Cref{fig:dp-cifar} shows the progress of the existing DP-NAS and AdvantageNAS on the same experiment as that in \Cref{ex:cifar}.
The architecture for it is obtained by \texttt{argmax}.
AdvantageNAS achieves a computational time efficiency comparable to that of SNAS and outperforms DARTS and NASP.

\section{Convergence Analysis by Entropy of Distribution} \label{apdx:entropy}
\begin{figure}[t]
  \centering
  \includegraphics{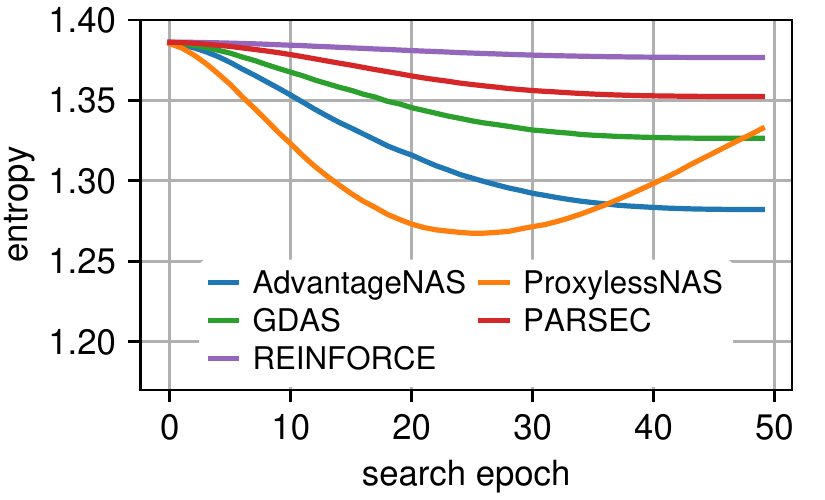}
  \caption{Transition of the entropy of distribution in the CIFAR-10 search process. The lines represent the averages over three search trials.}\label{cifar-10-entropy}
\end{figure}

We claim that AdvantageNAS converges faster than the existing SP-NAS.
To assess the convergence speeds of SP-NAS, we analyze the transition of the entropy of architecture distribution.
Specifically, in \Cref{cifar-10-entropy}, we show the progress of $\frac{1}{|\mathcal{I}|} \sum_{i \in \mathcal{I}} \mathcal{H}(\mu_i(\theta_i))$, where $\mathcal{H}$ is a function that returns the entropy of the distribution.
In our problem setting, the goal of NAS is to select one operation from the candidate operations.
Hence, if the search process converges, then the categorical distribution over candidate operations seems to become a Dirac delta after the entropy of categorical distribution decreases.
We observe that the entropy decreases more quickly in \anas{} than it does in REINFORCE, GDAS, and PARSEC.
In the initial steps of the search process, the entropy of ProxylessNAS decreases faster than that of AdvantageNAS.
However, the entropy of ProxylessNAS increases after around 25 epochs and thus does not converge.

\section{Visualization of Search Result} \label{vis-arc}
 
\begin{figure}[b]
  \centering
  \includegraphics[width=0.6\hsize]{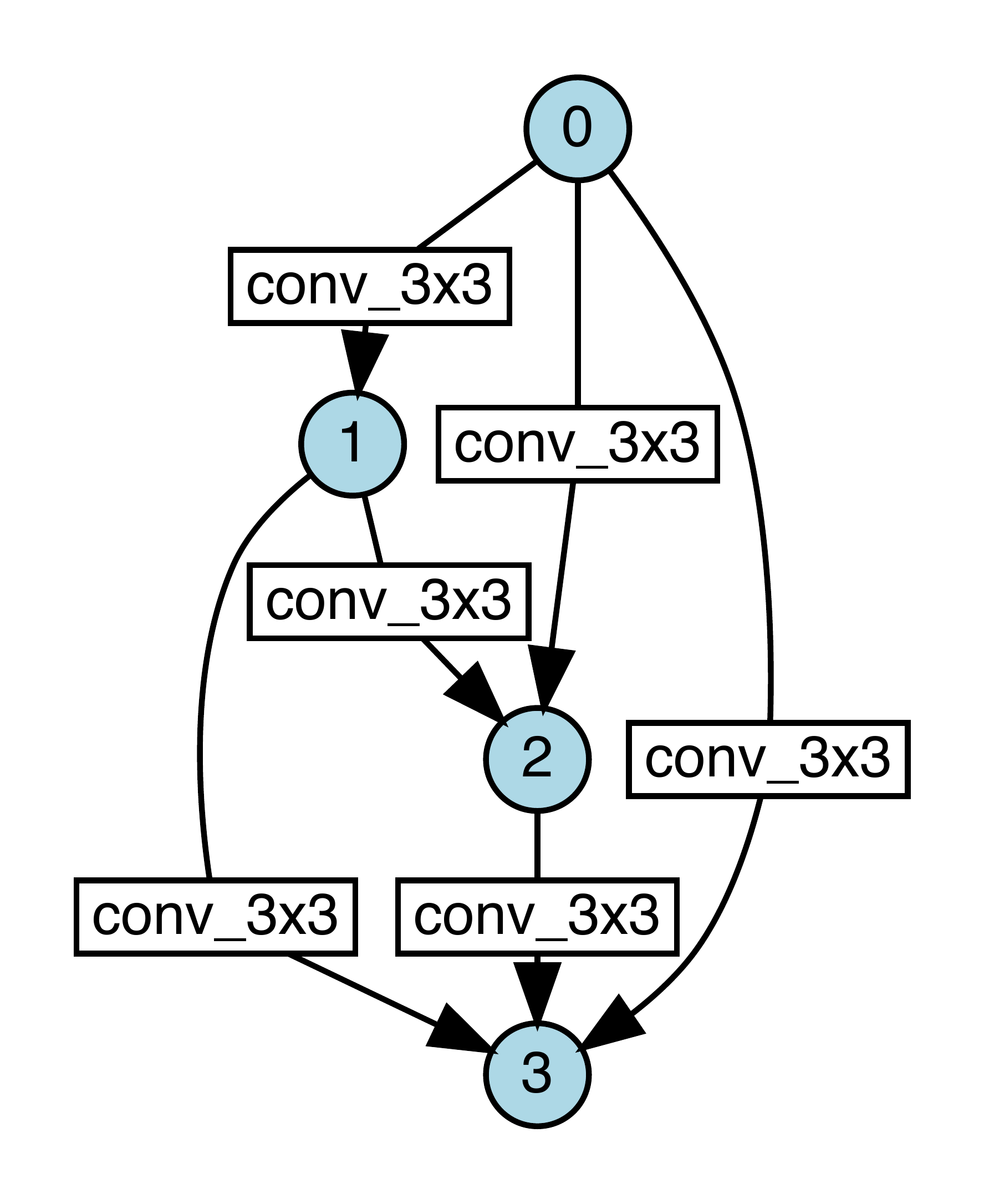}
  \caption{Architecture obtained by AdvantageNAS on CIFAR-10. It is the best architecture in terms of the test accuracy over three trials. It achieved a test accuracy of $93.76$.}
  \label{arc-cifar-a}
\end{figure}

\begin{figure}[b]
  \centering
  \begin{subfigure}{0.48\hsize}%
  \includegraphics[width=\hsize]{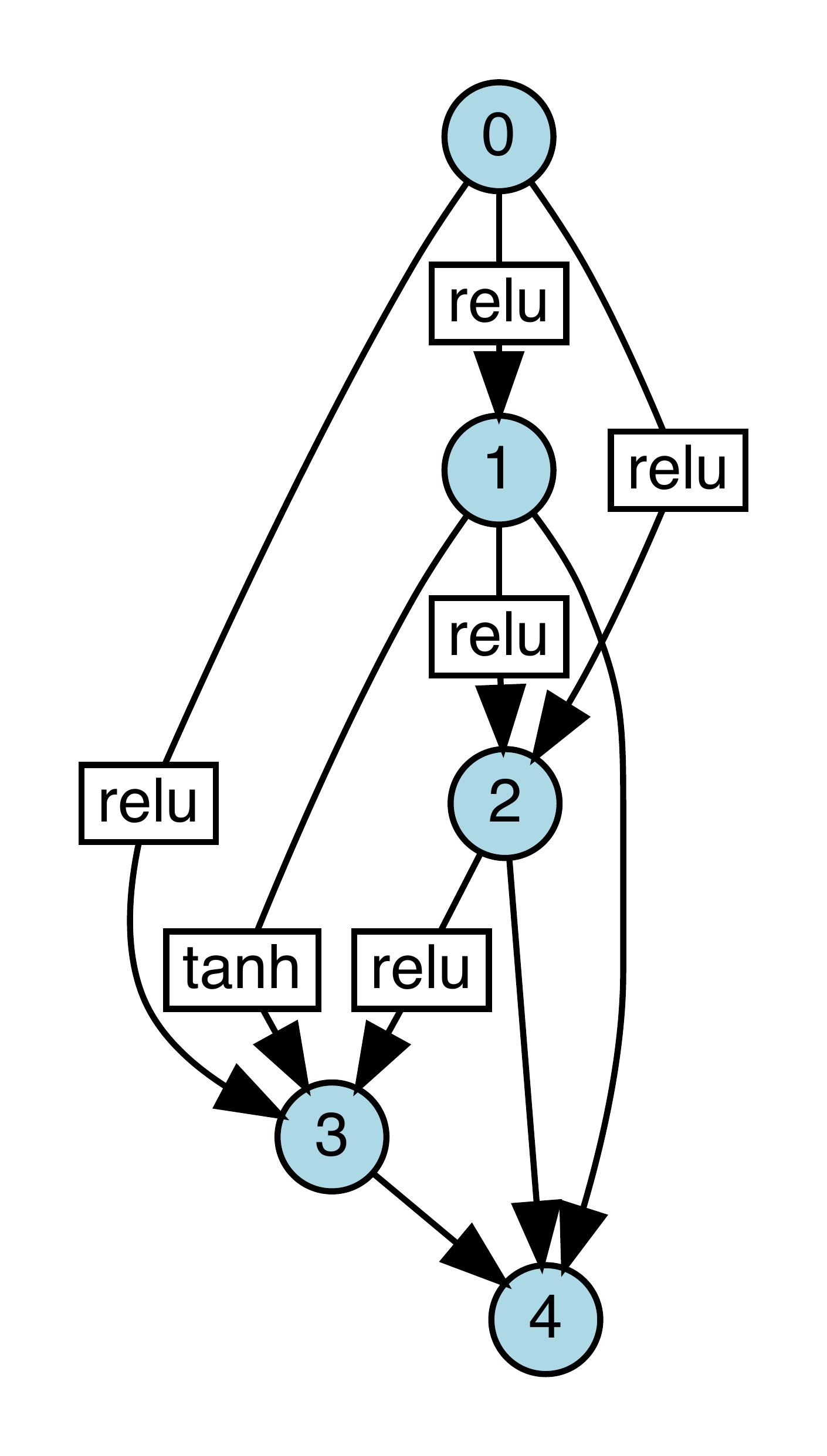}%
  \caption{AdvantageNAS on PTB}\label{arc-ptb-a}%
\end{subfigure}%
\begin{subfigure}{0.48\hsize}%
  \includegraphics[width=\hsize]{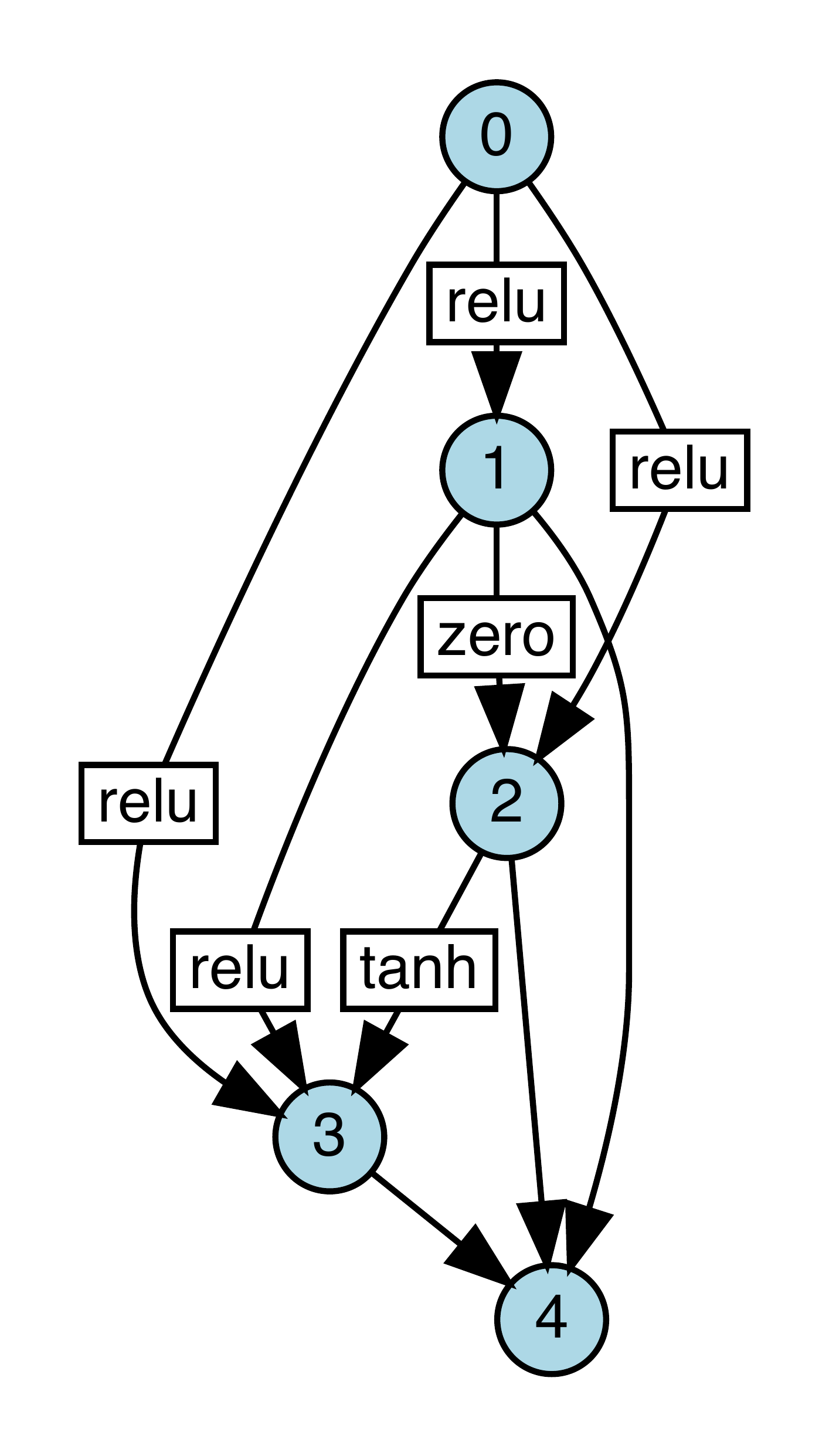}%
  \caption{ProxylessNAS on PTB}\label{arc-ptb-p}%
\end{subfigure}%
\caption{Architecture obtained on PTB.
  (a) The best architecture obtained by AdvantageNAS in terms of validation perplexity over five trials. It achieved a validation perplexity of $94.66$ and test perplexity of $91.12$.
  (b) The best architecture obtained by ProxylessNAS in terms of test perplexity over five trials. It achieved a validation perplexity of $95.00$ and test perplexity of $90.42$.}
\end{figure}

\Cref{arc-cifar-a,arc-ptb-a,arc-ptb-p} shows some architectures obtained in our experiments.
The circles correspond to nodes and the number in each circle denotes the index of the node.

\end{document}